# TransMatch: A Transfer-Learning Framework for Defect Detection in Laser Powder Bed Fusion Additive Manufacturing

Mohsen Asghari Ilani, Yaser Mike Banad

School of Electrical and Computer Engineering, University of Oklahoma, Norman, 73019, U.S.A. (e-mail: bana@ou.edu).

**Abstract**
Surface defects in Laser Powder Bed Fusion (LPBF) pose significant risks to the structural integrity of additively manufactured components. This paper introduces TransMatch, a novel framework that merges transfer learning and semi-supervised few-shot learning to address the scarcity of labeled AM defect data. By effectively leveraging both labeled and unlabeled novel-class images, TransMatch circumvents the limitations of previous meta-learning approaches. Experimental evaluations on a Surface Defects dataset of 8,284 images demonstrate the efficacy of TransMatch, achieving 98.91% accuracy with minimal loss, alongside high precision, recall, and F1-scores for multiple defect classes. These findings underscore its robustness in accurately identifying diverse defects, such as cracks, pinholes, holes, and spatter. TransMatch thus represents a significant leap forward in additive manufacturing defect detection, offering a practical and scalable solution for quality assurance and reliability across a wide range of industrial applications.

## 1. Introduction

Laser Powder Bed Fusion (LPBF), an advanced additive manufacturing (AM) technique, offers an innovative method for constructing intricate metallic components directly from computer-aided design (CAD) files. This advanced process involves the meticulous melting of alloy powder in a layer-by-layer fashion, as referenced in several studies [1–3]. In the LPBF process, a recoater mechanism deposits a thin layer of alloy powder onto a build plate. Subsequently, a laser is employed to selectively melt this powder layer, meticulously tracing the part model's specified geometry. Upon the completion of one layer, the build platform lowers, allowing the addition of a new powder layer. This cycle of layer deposition and laser melting continues in a sequential manner, facilitating the creation of the entire three-dimensional object, one layer at a time.

The process, directly linked to thermal dynamics, is prone to instability and heat effects from rapid cooling and melting. This can cause internal and surface defects in AM-fabricated surfaces [4–6]. Addressing such defects is critical, especially in applications demanding high structural integrity and resistance to cyclic loading. Overlooking these defects could impair the performance of AM components during operational use. Therefore, understanding defect formation mechanisms in fusion-based processes is vital for establishing precise process parameters tailored to the specific alloy system and chosen processing technique.

AM processes are broadly categorized into powder-bed fusion (PBF), including Selective Laser Melting (SLM)[7, 8] and Selective Laser Sintering (SLS)[9] and Directed Energy Deposition (DED)[10]. PBF spreads a thin powder layer on a build platform, where a focused energy source (laser or electron beam) selectively fuses powders, melting or sintering them onto a substrate, creating complex, net-shaped components.

Similar to heat-affected manufacturing processes like welding[11] and Electro-discharge Machining (EDM)[12, 13], AM processes have common physical characteristics, such as a mobile heat source, as observed in SLM (***Figure 3***). The formation of a fusion zone with recirculating liquid metal underlines the complexity of transient conditions. These conditions affect the metallurgical quality, microstructures, residual stresses, and distortions [6]. Processing instabilities and phenomena like Marangoni and Buoyancy forces contribute to superficial and internal defects, including keyhole porosity- pinholes and holes, spattering, Pore, and Lack-of-Fusion (LoF) (as illustrated in ***Figure 1***).

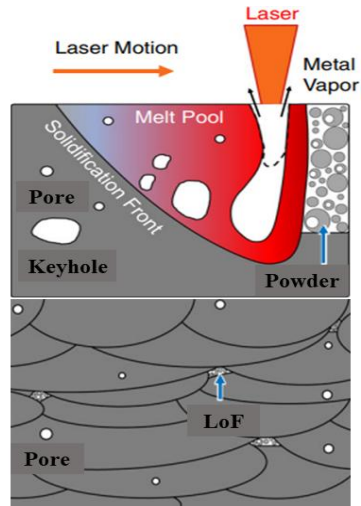

**Figure 1.** AM-built Surface Defects [14].

**1.1 Defect Formation Mechanisms and Mitigation Strategies**

Additive Manufacturing technologies, such as LPBF, offer numerous advantages compared to traditional manufacturing processes. They can fabricate intricate parts with complex shapes, consolidate assemblies into integral components, reduce lead time, and enable manufacturing in remote locations. However, AM parts are susceptible to volumetric defects[14]. In their machined condition, these defects—commonly referred to as lack of fusions (LoFs)[15], gas-entrapped pores[16], keyholes[16], and spatter—act as stress risers, adversely affecting the mechanical properties, especially fatigue resistance, of AM parts. In the case of LPBF processes, LoFs primarily arise due to insufficient overlap between adjacent melt pools, layers, or laser tracks, a consequence of inadequate energy input or excessively large hatch distance.

Conversely, gas porosity and entrapped pores manifest as spherical voids and occur due to gas trapped in raw metal powder particles or environmental inert gas during the melting process. The causes of gas pores are still under discussion, with Ng et al. [34] attributing their presence to pores within gas-atomized metal powder particles. Unlike LoF porosity, eliminating gas porosity is more challenging, with reported rates as high as 0.7%. Pores in gas-atomized powders are visible in cross-sectional optical images of powder particles, and the resulting pores in a laser deposit are depicted in *Figure 1*. In this study, our image spectroscopy methodology captured surface defects using Field Emission Scanning Electron Microscopes (FE-SEM), with a specific focus on superficial defects. Defects in AM-built components typically fall into three main categories: cracks (*Figure 2(a)*, porosity (holes (*Figure 2(b)*) and pinholes(*Figure 2(c)*), Marangoni-driven flow spattering (spatter) (*Figure 2(d)*).

**1.1.1 Crack**

Due to the rapid and repetitive cycles of heating, melting, and solidification induced by the moving heat source, different sections of the build undergo continuous thermal cycles. The dynamic interplay between the heat source and the feedstock results in a gradual material buildup, creating a history of thermal cycles and transient alterations in the geometry of the Additive Manufacturing (AM) component. This inhomogeneous growth direction subsequently leads to defects like cracks [17].

Solidification cracking in AM metals is a complex occurrence concentrated in the fusion zone near the end of solidification. This type of cracking, also known as hot cracking, is initiated by the interaction between metallurgical and mechanical factors, driven by a temperature gradient. Microstructures are influenced by thermal-metallurgical interactions, such as phase transformations, while local stress and strain behaviors are governed by thermal-mechanical interactions [18].

For solidification cracking to manifest, a combination of mechanical restraint (strain) and a susceptible microstructure must be present. *Figure 2(a)* outlines thermal-metallurgical factors contributing to solidification cracking in welding practices, which are also applicable to AM fusion-based processes. In specific alloy systems, cracks may form during the terminal stages of solidification, as evidenced by Reichardt et al.[19]'s attempt to transition from AISI 304L stainless steel(s) to Ti-6Al-4V, where the build process faced interruptions due to crack development.

Similar instances of cracks in deposited material before analysis have been documented by various researchers [5, 20]. These cracks initiate due to an accumulation of shrinkage strains along grain boundaries and interdendritic regions where a liquid film is distributed [6].

Primary material factors contributing to solidification cracking include the solidification temperature range (also known as the brittle temperature range) and the interfacial liquid morphology at terminal solidification stages. Alloys with a wide solidification temperature range are more prone to solidification cracking due to accumulated thermal strain proportional to the temperature range. Wide temperature ranges are often consequences of compositional variations leading to lower eutectics [21].

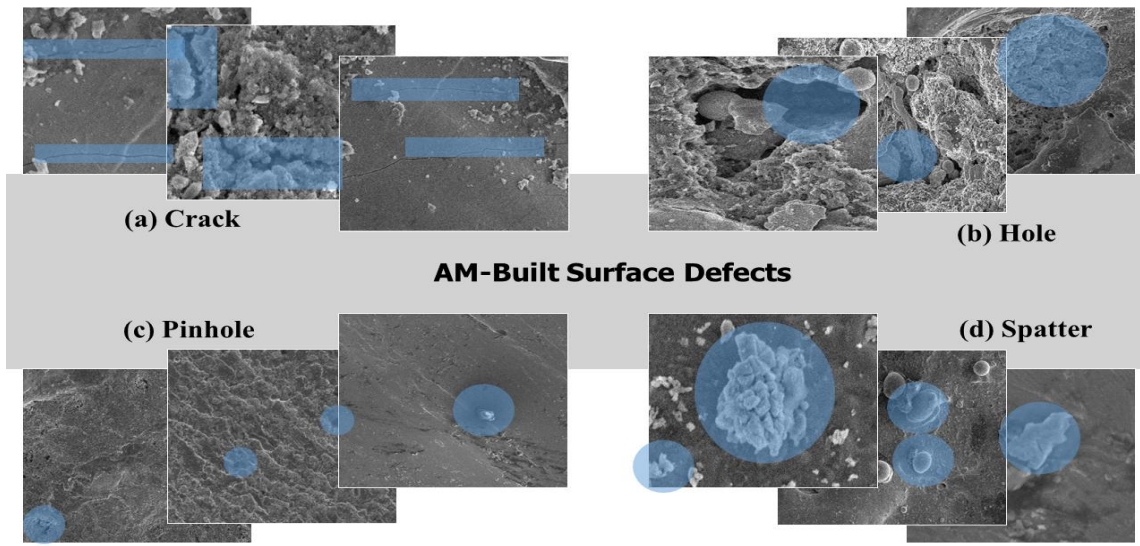

**Figure 2.** FE-SEM image of AM-built Surface Defects. (a) Crack, (b) Hole, (c) Pinhole, and (d) Spatter.

### 1.1.2 Pores (Pinhole and Holes)

Gas-entrapped and Keyhole pores are typically spherical in shape, as indicated in *Figure 2(b)* and *2(c)*. In PBF processes, these defects usually range in size from 5 to 20 μm [6], whereas in parts produced using DED, they are generally larger, often exceeding 50 μm. The formation and evolution of these pores are governed by processes such as gas entrapment, supersaturation of dissolved gases, and chemical reactions that generate gaseous species within the molten pool. The likelihood of nucleating gas-entrapped pores is higher when the equilibrium pressure of a gas within the molten pool exceeds the combined hydrostatic, atmospheric, and capillary pressures. Once nucleated, these pores serve as sites for the accumulation of vacancies, where supersaturated gases within a molten pool can diffuse [22]. Rapid cooling of the molten pool leads to the entrapment of these nucleation sites, whereas slower cooling and solidification rates allow the growth of these pores, and sometimes even their coalescence with adjacent pores. Once these pores reach a critical size, they detach from the solidification front and rise to the surface of the solidifying pool. It is termed a "Hole", *Figure 2(c)*, when the pore size exceeds 5, and conversely, it is referred to as a "Pinhole", *Figure 2(b)*, if the size is 5 or less [23].

Previous investigations have demonstrated that high solidification rates and high gas content in metal alloys can lead to increased concentrations of pores. Additionally, the concentrations of gases retained within each bubble plays a critical role in the growth of these pores, based on the imbalance in mass transfer between two fluids present in the Marangoni flow [5]. Gas porosity, resulting from gas entrapment within the powder feed system present in the powder particles, has emerged as a contributing factor to these crack formations. This type of porosity can appear at specific locations and typically exhibits a nearly spherical shape. Gas porosities are deemed highly undesirable, especially in terms of fatigue resistance and mechanical strength.

Keyhole porosity is a major concern in high-energy-density laser processes [24] and has been associated with macro porosity observed in laser and laser-arc hybrid welding techniques [25]. Insufficient control over keyhole-mode melting can result in the formation of unstable keyholes, which repeatedly form and collapse, leaving voids filled with entrapped vapor within the deposit. The primary mechanism behind keyhole porosity involves the creation of a deep V-shaped melt pool and the vaporization of elements within the pool [11].

In PBF process, keyhole porosity arises from the rapid, local melting of powder by a continuous laser heat source. The various stages of keyhole formation can be effectively illustrated using dynamic X-ray imaging. The impact of the laser on the build surface triggers oscillatory behavior that disperses the molten pool across the surface. Occasionally, instabilities in these molten pools can lead to the ejection of material, resulting in a pore forming at the bottom of the pool. This pore is then trapped and solidifies beneath the surface. Keyhole pores vary in their shape and typically range in size from 10 to 50 μm. Pores that are not connected to the surface can potentially be healed through post-processing techniques such as Hot Isostatic Pressing (HIP) [18].

### 1.1.3 Marangoni-Driven Flow Spattering

Marangoni-driven flow plays a significant role in the powder bed fusion (PBF) process, particularly in Laser Powder Bed Fusion (LPBF), influencing the melt pool dynamics and, subsequently, contributing to the occurrence of spatter. In PBF processes, precise control over the behavior of the melt pool is essential for maintaining the desired surface quality and dimensional accuracy of the manufactured parts. Marangoni-driven flow is a phenomenon driven by surface tension gradients within the melt pool. The temperature gradient across the liquid surface causes variations in surface tension, leading to fluid motion. In LPBF, this flow is induced by the localized heating from the laser, resulting in the migration of molten material toward regions with lower surface tension. As the Marangoni-driven flow occurs, it can cause disruptions within the melt pool, leading to the ejection of molten material in the form of spatter. The flow instabilities induced by the varying surface tension can cause the molten material to be expelled from the melt pool, resulting in undesirable droplets or splashes. This phenomenon can contribute to defects in the final manufactured parts.

Understanding the spattering mechanism is pivotal for its effective mitigation, particularly on heterogeneous surfaces. Young et al.[26] investigated spattering on a homogeneous powder bed using *in-situ* high-speed, high-energy X-ray imaging. In the context of laser powder bed fusion, spattering is classified into five styles, as determined by the laser power versus scanning speed map: solid spatter, metallic jet spatter, powder agglomeration spatters, entrainment melting spatters, and defect-induced spatters. Although this work primarily focused on the longitudinal direction of the melt pool, their classification and mechanistic explanations offer valuable insights into spattering occurring in the transverse direction of the melt pool, as shown in ***Figure 2(d)***.

### 1.2 Machine Learning for Defect Detection in Metal Additive Manufacturing

The process of Additive Manufacturing is highly complex, and the formation of parts within this domain is significantly influenced by various control parameters such as laser power, scanning speed, powder feed rate, and the quality of the powder feedstock. Traditionally, researchers have focused on optimizing these process parameters to enhance AM-built performance, and thereby, mitigate defects [27]. However, the manual adjustment of these diverse parameters to identify the optimal process setting is a labor-intensive and time-consuming endeavor. A promising solution to streamline this process involves leveraging machine learning algorithms for automatic defect detection [27–30]. For example, Barua et al. [31] utilized melt-pool gradient temperature deviation for predictive gas porosity detection in the laser additive manufacturing process. This approach not only promises to enhance the efficiency and effectiveness of the manufacturing process but also contributes to a more consistent and predictable quality in the final products.

One of the significant challenges in metal additive manufacturing is surface defects, which can adversely affect the mechanical properties of each deposition layer. This has prompted the development of machine learning (ML) based inspection methods. These evolving AI-driven techniques aim to provide more effective defect detection in AM processes. ML methods for defect detection can be broadly categorized into three learning principles: supervised, unsupervised, and semi-supervised learning.

Supervised learning involves training the algorithm with a labeled dataset, where the model learns to map input data to corresponding output labels, such as classifying images as defective or non-defective. Unsupervised learning, on the other hand, involves training on unlabeled datasets, where the algorithm must identify patterns without explicit guidance. This is particularly useful when labeled data is scarce. Semi-supervised learning merges elements of both, utilizing a mix of labeled and unlabeled data, offering a more versatile and scalable solution. While semi-supervised deep learning has not been widely applied in defect inspection for automated manufacturing, it has shown promising results in various fields like image processing, object recognition, and natural language processing

The complexity of the AM process often challenges traditional analytical methods in accurately predicting outcomes. Convolutional Neural Networks (CNNs), accelerated by Graphics Processing Units (GPUs), are emerging as a powerful tool for real-time monitoring, quality inspection, and evaluation in AM. The field saw significant advancements with LeCun et al. [32]'s introduction of the LeNet model, using convolution and pooling layers for digit recognition. Despite its innovation, the advancement of CNNs initially encountered hurdles due to computational performance constraints. This obstacle was significantly mitigated when Krizhevsky et al. [33] introduced the AlexNet model, marking substantial improvements in both accuracy and computational efficiency for ImageNet dataset classification using CNNs. The integration of CNN-based methods in AM process represents a substantial leap forward, providing a highly promising strategy for efficient and accurate real-time monitoring and quality inspection of additive build parts [34, 35]. Another key area of development is consistency regularization techniques in semi-supervised learning, which aim for consistent output predictions across input variations. Techniques like dropout and random data augmentations are used for creating perturbed inputs. MixMatch and temporal ensembling are examples of this, employing data augmentation methods like Mixup for consistency regularization. However, generating domain-specific data augmentations can be challenging.

Among the various categories of semi-supervised learning approaches, consistency regularization and pseudo-labeling stand out as prominent techniques [36, 37]. Pseudo-labeling is a technique that enhances labeled training data by automatically labeling unlabeled data and incorporating them into the original training set. The process involves training a model using only labeled data, predicting labels (pseudo-labels) for unlabeled data, and then training a new or updated model using the combination of original labeled and pseudo-labeled data. The quality of pseudo-labels is often considered more important than their quantity. Different approaches have been proposed for generating and selecting pseudo-labels, with considerations for confidence levels. Lee et al. [38], for example, calculates pseudo-labels using a single network trained on labeled data, updating the network with all pseudo-labeled data regardless of confidence. However, training with all pseudo-labels can lead to noise, prompting the need for considering only high-confidence pseudo-labels. Various soft-thresholding approaches have been proposed, including those based on prediction confidence and information entropy. Shi et al. [39] additionally consider local neighborhood density for outlier consideration.

Recent advancements include an uncertainty-aware pseudo-labeling selection method has been proposed [38], demonstrating improved performance over various pseudo-labeling techniques. This method computes the uncertainty estimation of an image based on predictions obtained from the network for that specific image [40, 41]. This highlights the ongoing efforts to refine and enhance the effectiveness of semi-supervised learning techniques in diverse applications. Our study introduces TransMatch, a novel approach tackling the challenge of extensive labeling in big data scenarios. TransMatch employs four steps pseudo-labeling for supervised learning-based CNN models, combined with transfer learning for unsupervised or unlabeled images. Here, features from a pre-trained CNN model are applied to vast amounts of unlabeled data, incorporating elements of few-shot learning. This semi-supervised learning (SSL) approach is particularly effective in making predictions on an unlabeled series of images based on a small set of labeled images, thus addressing some of the major challenges in AM defect detection.

## 2. Research Methodology

In the research methodology, the selective utilization of powder melting techniques plays a pivotal role. Specifically, the process of Selective Laser Melting (SLM) involves the complete melting of powders, as illustrated in *Figure 3 (a)*. This method, implemented in Laser Powder Bed Fusion (LPBF), is characterized by the full liquefaction of the powder material. On the contrary, Selective Laser Sintering (SLS) comes into play when dealing with sintered powders, where particles undergo partial melting.

In the context of our methodology, the distinction between SLM and SLS is particularly relevant. SLM, operating with fully melted powders, is instrumental in achieving precise fabrication within LPBF processes. Meanwhile, SLS, which involves the partial melting of powders, serves a distinct purpose in our research. Specifically, SLS is strategically employed for monitoring surface defects during the ex-situ monitoring stage. This careful selection of powder melting techniques and their strategic application aligns with the research objectives, contributing to the comprehensive evaluation of surface defects in the additive manufacturing process. The distinct attributes of SLM and SLS are harnessed to optimize our approach in investigating and analyzing surface defects, enhancing the overall reliability and accuracy of our research outcomes.

This study utilizes datasets consisting of images captured through Field Emission Scanning Electron Microscopy (FE-SEM) in ex-situ monitoring stage, as shown in *Figure 3(b)*. This imaging technique is typically conducted in a high vacuum environment to minimize disruptions from gas molecules, which could interfere with the electron beam and the emitted secondary and backscattered electrons used for imaging. A total of 232 topographical images, each with a resolution of 1536×1103 pixels (1536 pixels in width and 1103 pixels in height), are analyzed, as depicted in *Figure 3 (b)*. Furthermore, the labeling of our dataset, comprising 34 labels, into the four defect classes (Crack, Pinhole, Hole, and Spatter) is detailed in *Figure 3 (c).* This labeling process is facilitated by a novel annotated methodology, providing a structured and comprehensive classification for further analysis.

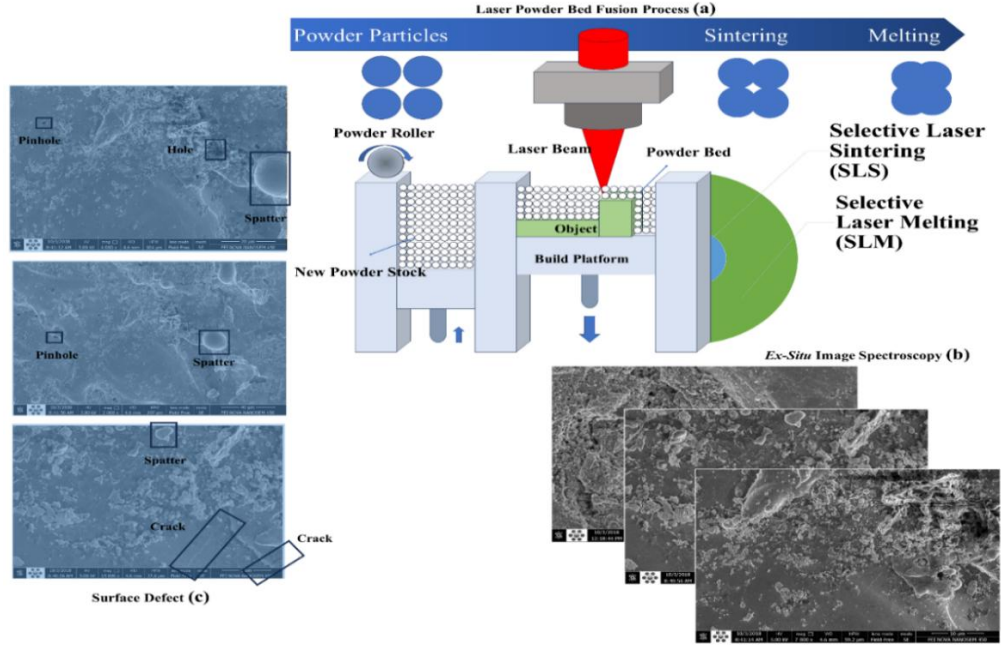

**Figure 3**. *In-Situ* and Ex-Situ Monitoring of LPBF process. (a) Diagram illustrating Fusion-Based Additive Manufacturing Process, (b) FE-SEM image of AM-built Surface Defects in Ex-Situ Monitoring Stage, and (c) Monitoring Surface Defects Monitoring.

### 2.1 Image Preprocessing

In this section, we elucidate the comprehensive data preprocessing pipeline employed in our study, which encompasses both supervised and semi-supervised learning approaches. Our dataset consists of two categories: labeled and unlabeled images. The labeled images, crucial for supervised learning, were meticulously annotated using the LabelImg method, resulting in 34 labeled images. From these images, 14984 annotations were extracted, providing a detailed understanding of the images' content through bounding boxes and labels.

For supervised learning, we utilized a CNN model to train and test on the labeled images. The dataset was split into two subsets: one for training and another for testing. The training subset comprises 6,742 images for the supervised learning model (***Figure 4 (b) and 4(c)***) and 7,455 images for the unsupervised learning model (***Figure 4(d)*** and ***4(e)***), denoted as (x-train, y-train). The testing subset contains 750 images for the supervised learning model and 829 images for the unsupervised learning model, designated as (x-test, y-test). Notably, the dataset was imbalanced, prompting the incorporation of the compute-weight function from scikit-learn to balance the class distribution during model fitting.

Before proceeding with both supervised and semi-supervised learning, an essential step involved the application of image preprocessing techniques using the OpenCV library. The objective was to enhance the detection of defect edges within the images. The sequential preprocessing steps are outlined as follows, as shown in ***Figure 4 (a)***:

  I.  Conversion to Grayscale (COLOR_BGR2GRAY): The RGB images were converted to grayscale, simplifying subsequent processing while retaining crucial structural information.
 II.  Gaussian Blur: To reduce noise and emphasize prominent features, we applied a Gaussian blur to the grayscale images.
III.  Fast Non-Local Means Denoising: This approach was employed to further refine the images and enhance the clarity of defect boundaries.
 IV.  Adaptive Thresholding: This technique was utilized to segment the images, enhancing the contrast between defect and non-defect regions.
  V.  Edge Detection (Canny): Finally, the Canny edge detection algorithm was applied to identify edges within the images, aiding in the precise localization of defects.

These preprocessing steps collectively contribute to the creation of a refined dataset, optimized for both supervised and semi-supervised learning methodologies. The resultant images, as illustrated in ***Figure 4***, showcase the effectiveness of the preprocessing techniques in accentuating defect boundaries and facilitating subsequent model training and evaluation.

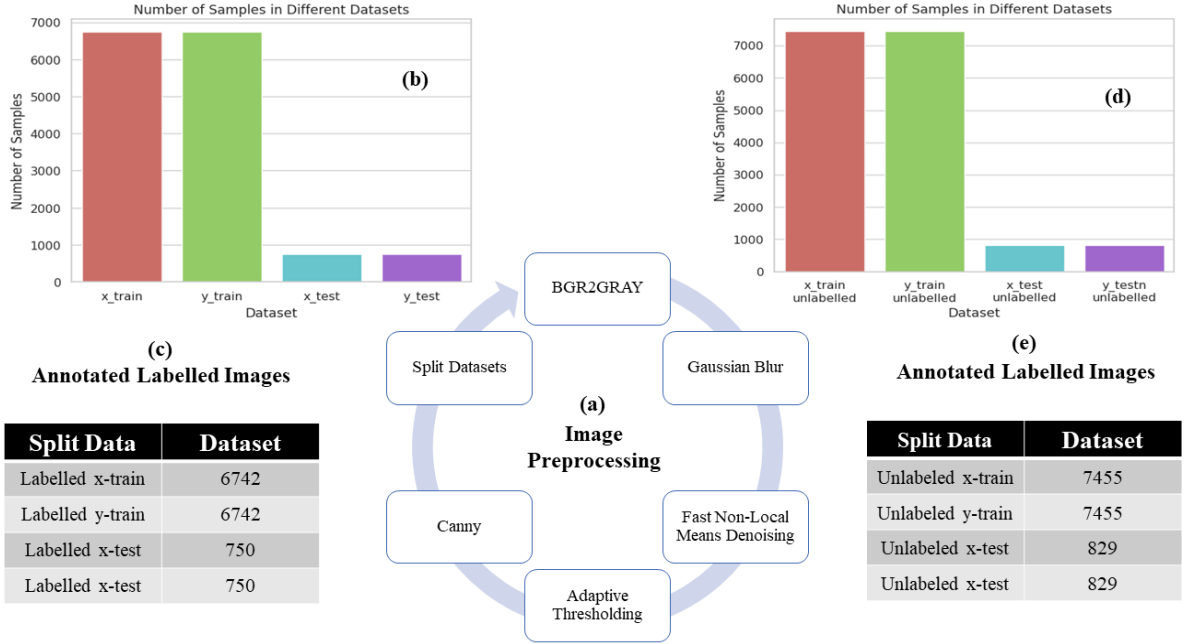

**Figure 4**. Pre-processing Procedures for Labelled and Unlabeled Data in Supervised and Semi-Supervised Learning Approaches. (a) Pre-processing for both Labelled and Unlabeled Images, (b) Division of Pre-processed Labelled Images, (c) Numerical Data of the Divided Labelled Images, (d) Division of Pre-processed Unlabeled Images, and (e) Numerical Data of the Divided Unlabeled Images.

## 2.2 Supervised Learning of Object Detection

### 2.2.1 Pre-Trained Convolutional Neural Networks (CNNs) Model

The incorporation of Convolutional Neural Networks [90] in the context of AM-built surface defect detection unfolds through two primary scenarios. The first involves crafting a sophisticated, multi-layer CNN structure, extracting image features from a separate network, and subsequently conducting end-to-end training for image defect detection [42–44]. In contrast, the second scenario integrates CNN with an additional training model, leveraging results from the pre-trained stage, denoted as semi-supervised learning. This section distinctly focuses on the application of a supervised CNN model to labeled images annotated using the LabelImg framework, as illustrated in ***Figure 5***.

CNNs demonstrate prowess in learning high-dimensional data and extracting abstract, essential, and high-level features. However, deeper network structures can escalate computational complexity, potentially leading to confusion and overfitting during the training, validation, and testing phases. Recognizing these challenges, numerous studies have delved into defects arising from Laser Powder Bed Fusion, spanning both the surface and interior of AM-built parts.

The process of manually labeling images for extensive datasets proves challenging, time-consuming, and susceptible to oversights. To address these concerns, an initial application of a CNN model focuses on a subset of labeled datasets, capturing feature weights associated with surface defects—Crack, Pinhole, Hole, and Spatter. Subsequently, these weights are transferred under transfer learning for broader applicability. Images obtained from the LPBF-built surface using the Field Emission Scanning Electron Microscopy (FE-SEM) technique pose a labeling challenge due to their intricate nature. Labeling is accomplished using the LabelImg tool (***Figure 5 (a)***), and objects are extracted based on the bounding box coordinates and label names obtained from labelled image annotations. Subsequent preprocessing steps (***Figure 4(a)***) are applied to prepare the data for training in the CNN model. The resultant dataset comprises 13,484 training sets and 1,500 test datasets (***Figure 4(d)***). The architecture of the CNN involves several layers, including convolutional and pooling layers, followed by flattening and dense layers for classification. A visual representation of the CNN architecture is provided in Figure 5, illustrating the model's performance on the datasets. ***Figure 5(b)*** showcases the high accuracy of the CNN model on labeled datasets (98% for Crack, 99.6% for Pinhole, 98% for Hole, and 98.7% for Spatter), confirming the effectiveness of the model and associated cleaning methodologies.

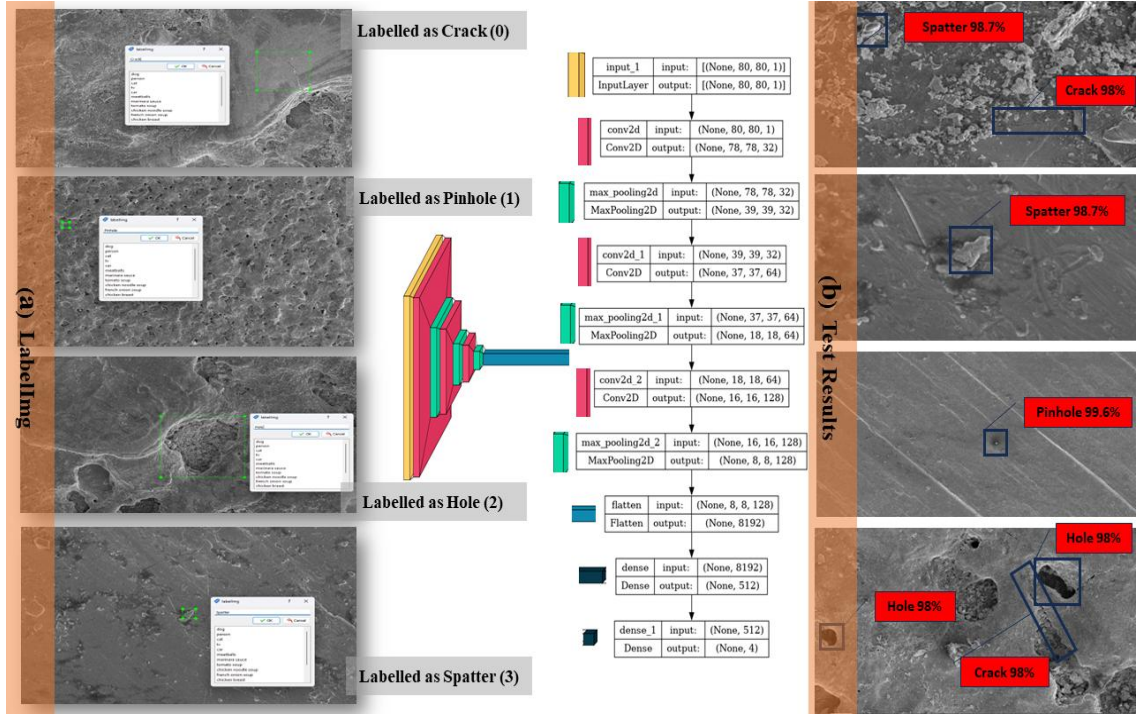

**Figure 5.** Supervised-CNN Learning Architecture of Labelled Images. (a) Labelling Images in LabelImg framework. (b) Test results accuracy.

### 2.2.2 Pseudo-labeling

In this study, we employ a unique approach involving iterative rounds of training a CNN for AM-built surface defect detection. The process unfolds in several steps within each round. Initially, a training set is established and labeled by human annotators using the LabelImg framework (*Figure 5(a)*). Subsequently, this initial training (*Figure 6(1-2)*) set is used to train a neural network specifically designed for defect detection (*Figure 6(3)*). The trained network then performs predictions on a larger unlabeled dataset (*Figure 6(5-7)*), extending its capabilities to identify defects in a broader context. In the final step of each round, the unlabeled images with predictions deemed confident, based on a predefined confidence threshold, are assimilated into the training set (*Figure 6(8-9)*). This expanded training set, now comprising both previously labeled and newly added unlabeled images, is utilized to train a new neural network for the subsequent round (*Figure 6(10-11)*).

The utilization of a confidence threshold is pivotal in governing the inclusion of unlabeled images in the training set. Only those images surpassing the specified confidence threshold are incorporated, serving as a trade-off between the size of the training set and the quality of pseudo-labels. This threshold value effectively determines the probability range at which defects are classified into four softmax categories. The Python snippet provided below exemplifies how this thresholding mechanism is implemented in our approach [46]. This iterative process ensures a dynamic balance between the growth of the training set and the refinement of pseudo-label quality throughout successive rounds of training.

The subsequent neural network is trained by incorporating both the initially manually labeled dataset and the subset of the unlabeled dataset where prediction probabilities exceeded the predetermined threshold. Following this training phase, a fresh set of predictions is executed on the remaining unlabeled data, and the newly acquired images along with their corresponding predictions are integrated into the subsequent pseudo-labeling iteration. This iterative cycle continues until a predefined stopping condition is met. In our study, we conducted up to four rounds of pseudo-labeling for each experiment. *Figure 6* provides a visual representation of the sequential steps constituting a single round of pseudo-labeling. This diagram succinctly illustrates the iterative nature of the process, where predictions from the previous round contribute to the expansion of the training set in subsequent rounds, enhancing the model's capacity to discern defects in the unlabeled dataset.It is crucial to highlight the distinction between the approach explored in this study and the one proposed by [38], which advocates for the simultaneous use of labeled and unlabeled data throughout the training process, with pseudo-labels being recalculated after every weight update. In contrast, our methodology involves multiple rounds of training, where new unlabeled data and corresponding predicted labels are incorporated only after completing full training schedules. This deliberate choice enables us to set a threshold value after each training round, ensuring that only unlabeled images with prediction probabilities exceeding 50% are included. This differs from the approach suggested by [38], which incorporates every unlabeled image throughout the entire training procedure without applying a threshold. Moreover, our method can seamlessly integrate with previously trained networks without necessitating modifications to the original network architecture or optimization procedures. This adaptability serves as a valuable additional step for enhancing existing defect identification networks.

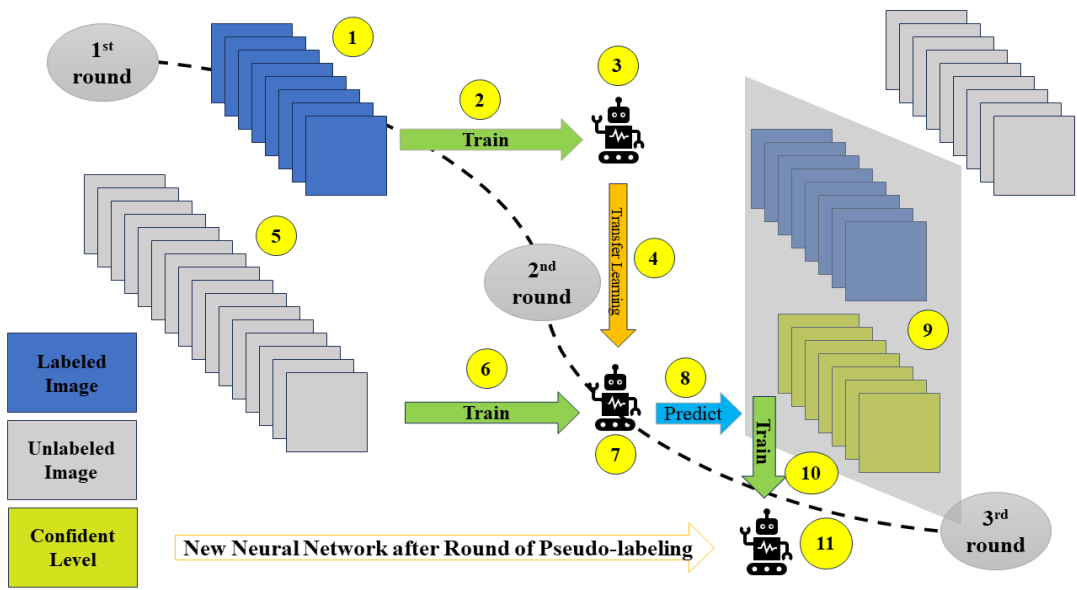

**Figure 6.** Pseudo-labeling round: initial network training (1-4), unlabeled data predictions (5-8), and new network training with confidently predicted labels (9-11). Blue points: labeled data, gray: unlabeled, orange (confidence above threshold): added to training set.

### 2.2.3. Semi-Supervised Learning (SSL)

In the landscape of deep learning, marked progress has been witnessed across various domains, albeit the persistent challenge of acquiring extensive labeled data. This paper addresses this challenge in the context of surface defect detection in Laser Powder Bed Fusion processes. Inspired by the human ability to learn from scarce examples, we propose TransMatch—a pioneering framework that merges Semi-Supervised Few-Shot Learning with a Transfer Learning approach, emphasizing the strategic use of pseudo-labels to enhance defect detection.

Qi et al. [45] introduced a method where the classifier weights for novel classes are imprinted using the mean vectors of feature embeddings from few-shot examples. Qiao et al. [46] devised a mapping function that learns the association between activations (i.e., feature embeddings) of novel class examples and classifier weights. Gidaris et al. [47] proposed an attention module to dynamically predict classifier weights for novel classes. Chen et al. [48] demonstrated that transfer-learning based approaches, such as the one proposed by Qi et al. [45]. In line with this concept, our proposed framework shares similarities with, involving the pretraining of a feature extractor and using it to extract features for few-shot examples from novel classes, subsequently imprinting classifier weights.

### 2.2.4. Transfer Learning

Transfer learning is a concept that involves employing a pretrained model, which has already undergone training on a substantial dataset, for a task similar to our target task. In the context of surface defect detection in our unlabeled images (*Figure 7(a)*), a pretrained model can be utilized to identify pertinent patterns and features associated with surface defects. The next step involves fine-tuning (*Figure 7(b)*) or adapting this pretrained model to the specific task of surface defect detection, achieved by training it on a more limited dataset containing images of defective surfaces. Fine-tuning encompasses the adjustment of the pretrained model's weights using the new dataset of images (*Figure 7(c)-(f)*). This process enables the model to glean insights from the new data, refining its internal representations to enhance surface defect detection capabilities. Leveraging transfer learning allows us to harness the pretrained model's adeptness in recognizing features and patterns, resulting in a significant reduction in the required volume of data and training time to construct a precise surface defect detection system. Transfer learning can be implemented using diverse pretrained models as different architecture of CNN (*Figure 7(e)*) such as VGG, ResNet, and Inception, typically trained on expansive image datasets like ImageNet, housing millions of images with a new approach of TansMatch algorithm (*Figure 7(g)*). Through transfer learning, we leverage the feature extraction capabilities inherent in these models, leading to a high level of accuracy and efficiency in surface defect detection. Subsequently, Loading the Processed Data into the TransMatch Algorithm (further depicted in the diagram).

To underscore the necessity of effective data utilization, we focus on extracting insights from both labeled base-class data and unlabeled novel-class data. This approach, termed semi-supervised few-shot learning, leverages readily available information from large-scale datasets for model pre-training and unlabeled data for new task adaptation.

Our framework encompasses three key components: (1) pre-training a model with a limited set of labeled data from AM-built surface classes; (2) employing the pre-trained model as a feature extractor for few-shot examples from novel classes; and (3) integrating state-of-the-art semi-supervised methods, such as TransMatch (*Figure 7(g)*)—a unique method that amalgamates transfer-learning and semi-supervised learning, to capitalize on unlabeled data.

Extensive experiments on labeled and unlabeled Surface Defects datasets showcase TransMatch's prowess in leveraging unlabeled data for few-shot learning, leading to state-of-the-art results. The initial labeled dataset, consisting of 34 images, is augmented through pseudo-labeling, enriching the training data for improved model performance. Implementation involves training a supervised CNN on the initially labeled data, followed by pseudo-labeling to augment the dataset. Subsequently, the TransMatch framework demonstrates the effectiveness of leveraging features and weights extracted from both labeled and unlabeled images using semi-supervised few-shot learning under transfer learning. As a goal of this study, our work introduces a novel paradigm for surface defect detection, combining semi-supervised few-shot learning and transfer learning under the TransMatch framework. The detailed experimental results underscore the potential of our approach to significantly advance defect detection, especially in scenarios with limited labeled data.

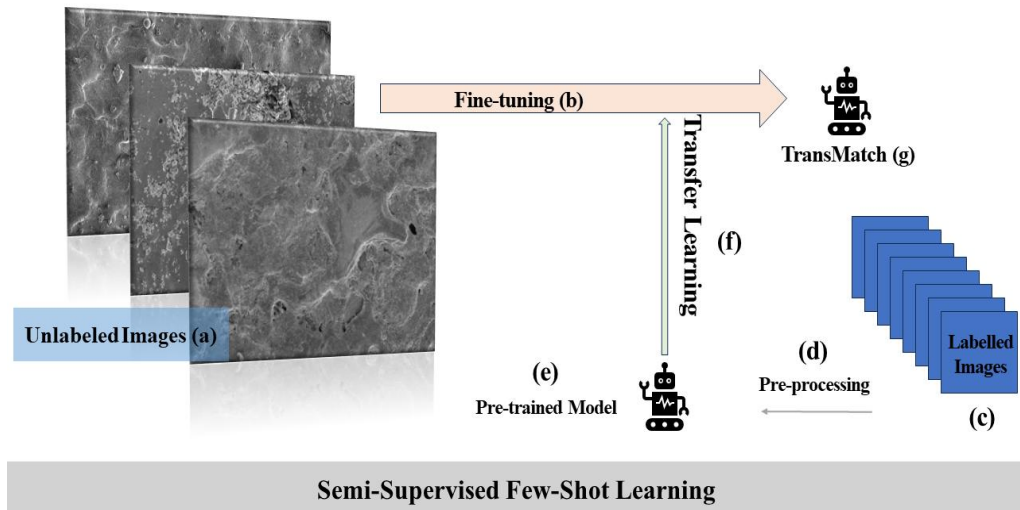

**Figure 7.** Semi-Supervised Few-Shot Learning (SSFSL) Architecture for Unlabelled Images Incorporating (a) Fine-Tuning (b), Feature Extraction via Pseudo Labeling (c), and Pre-processing (d) Utilizing a Pre-trained Model (e) Through Transfer Learning.

## 3. Results and Discussion

Semi-supervised learning, a powerful amalgamation of labeled and unlabeled data, provides an innovative solution to the challenges of handling limited labeled datasets. The methodology initiated with the manual labeling of 232 high-resolution LPBF images, distinguishing between Crack, Pinhole, Hole, and Spatter, using the LabelImg framework. This annotated dataset, although small, served as a foundational resource for subsequent model training (*Figure 4(a)*).

To amplify the efficiency and scale of our model, we leveraged the advantages of Semi-Supervised Few-Shot Learning (SSFSL) within the framework of transfer learning. The use of transfer learning (*Figure 6(4)*) facilitated the incorporation of knowledge from a pre-trained Convolutional Neural Network (CNN) model, which had been trained on a diverse dataset. This approach, well-established in computer vision tasks, allowed the model to harness generalized features from the pre-trained model, significantly boosting its ability to recognize subtle patterns and features related to LPBF defects.

The implementation of pseudo-labeling (*Figure 6(1-11)*) further contributed to the robustness of our approach. Pseudo-labeling involved using a partially trained model to predict labels for the larger, unlabeled dataset. Predictions exceeding a 50% confidence threshold were added to the labeled dataset during each iteration (*Figure 6(9)*). This iterative pseudo-labeling process helped gradually transform the unlabeled dataset into a more informative, labeled one, enabling the model to learn from a broader set of examples.

Transfer learning, coupled with pseudo-labeling, played a pivotal role in addressing the challenge of scarce labeled data by efficiently utilizing both labeled and unlabeled datasets. The model's ability to distill relevant information from the unlabeled dataset, with the guidance of the pre-trained model, contributed significantly to its overall accuracy.

The introduction of the TransMatch algorithm, a feature transfer mechanism, further augmented the model's learning capabilities. TransMatch facilitated the seamless transfer of features, ensuring that the knowledge gained during pseudo-labeling was efficiently integrated into the model. This strategic feature transfer played a crucial role in enhancing the model's discriminatory power, especially in distinguishing between defects with subtle visual differences.

With pseudo-labeling (*Figure 6(1-11)*), initially, 232 high-resolution images from LPBF were labeled individually for Crack (*Figure 2(a)*), Pinhole (*Figure 2(b)*), Hole (*Figure 2(c)*), and Spatter (*Figure 2(d)*) using the LabelImg framework (*Figure 5 (a)*). Preprocessing steps (*Figure 4 (a)*), including grayscale conversion, GaussianBlur, denoising, adaptiveThreshold, and Canny edge detection, were applied, resulting in a labeled dataset of 14,984 samples ( *Figure 4 (b)*).

Supervised CNN-based learning on labeled images achieved impressive results with 99% accuracy cross training, validation, and testing (*Error! Reference source not found.*). Precision, recall, and f1-score were consistently high for all defect types. To address the challenge of large datasets and time-consuming labeling, SSFSL was introduced. Transfer learning facilitated the transfer of feature weights from a pre-trained CNN model through four rounds of pseudo-labeling.

The results of the supervised Convolutional Neural Network (CNN)-based learning for images labeled by LabelImg are highly promising, demonstrating a robust performance in defect detection across various categories. The precision, recall, and f1-score metrics provide a comprehensive evaluation of the model's effectiveness (*Table 1*).

For defect Crack (Class 0), representing a specific defect category, the model achieved a precision of 1.00, indicating that when it predicted the presence of this defect, it was correct 100% of the time. The recall of 0.98 indicates that the model successfully identified 98% of the instances where this defect was present. The f1-score, a balanced measure of precision and recall, is 0.99, affirming the model's proficiency in capturing both false positives and false negatives. With the support of 50 instances, the model exhibited exceptional performance for this particular defect category.

Similarly, for defective Pinhole (Class 1), the model achieved high precision (0.99), recall (0.98), and f1-score (0.99). This suggests that the model accurately identified and classified instances of Class 1 defects, showcasing its reliability in this category. With the support of 124 instances, the model's accuracy in defect classification is commendable.

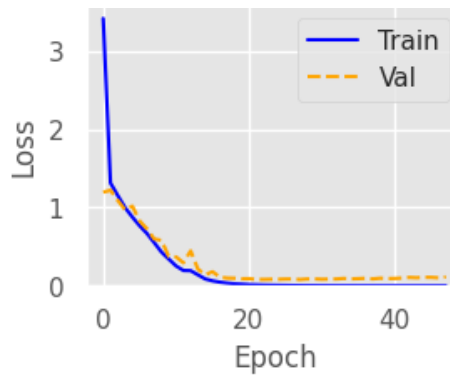

**Figure 8**. Pre-trained CNN-based training and validation.

Defect Hole (Class 2) is characterized by precision, recall, and f1-score values of 0.98, 1.00, and 0.99, respectively. These metrics indicate the model's near-perfect ability to discern and correctly classify instances of Class 2 defects. With 258 instances in the dataset, the model demonstrated remarkable accuracy for this defect category.

For defect Spatter (Class 3), the model achieved precision, recall, and f1-score values of 0.99, 0.98, and 0.99, respectively. The high precision suggests that the model rarely misclassifies instances as Class 3 when they belong to other categories. The recall of 0.98 indicates that the model effectively identified 98% of the instances of Class 3 defects. With 318 instances, the model exhibited strong performance in capturing instances of this defect type.

The overall accuracy of 99% across all defect types, as indicated by the macro and weighted averages, highlights the model's generalization capabilities. The weighted average takes into account the contribution of each defect type based on its support, providing a more balanced assessment of the model's overall performance. In summary, the supervised CNN-based learning approach, utilizing images labeled by LabelImg, achieved outstanding results, demonstrating its efficacy in defect detection for Laser Powder Bed Fusion processes.

Pseudo-labeling involved using a partially trained model to predict labels for the remaining unlabeled dataset. Confident predictions with over 50% confidence were added to the labeled dataset for each iteration (***Figure 6(9)***). The process is iterated through multiple rounds, enhancing the model's performance. The SSFSL approach, using TransMatch algorithm for feature transfer, demonstrated unprecedented outcomes for a substantial series of unlabeled images, achieving an overall accuracy of 98.91% (***Figure 9(a)***) and loss of 0.0188 (*Error! Reference source not found.*).

The Semi-Supervised Few-Shot Learning (SSFSL) approach with pseudo-labeling has proven to be a highly efficient method for defect detection in Laser Powder Bed Fusion processes. By leveraging a limited labeled dataset and iteratively incorporating pseudo-labeled data from a larger, unlabeled dataset, the SSFSL approach achieved remarkable results, showcasing its effectiveness in comparison to the supervised approach.

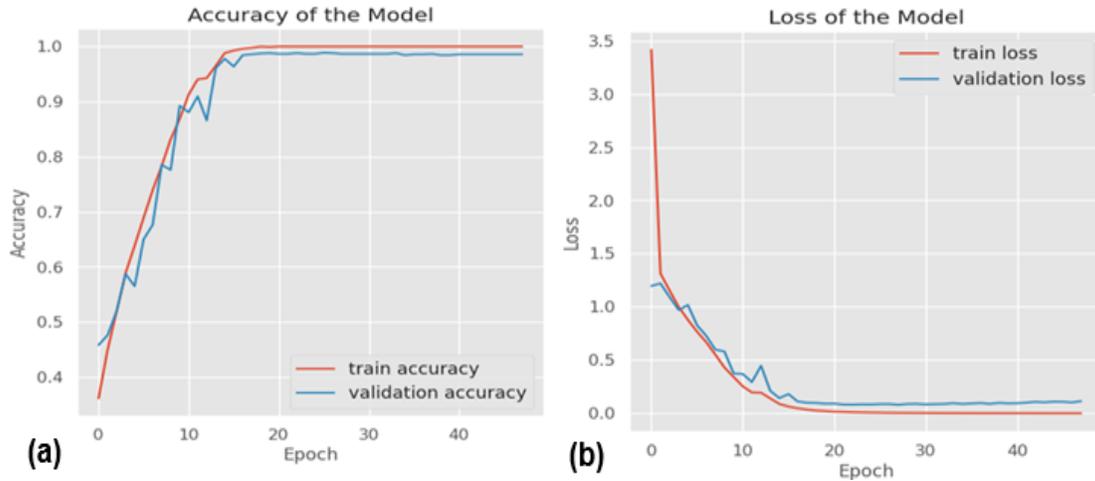

**Figure 9**. (a) Accuracy of SSFSL of unlabeled images (accuracy: 0.9891). (b) Loss error of SSFSL of unlabeled images (loss: 0.0188).

**Table 1**. Evaluation of Supervised CNN-based Model for images labelled by LabelImg.

| Label | Precision | Recall | F1-score | Support |
|---|---|---|---|---|
| Crack | 1.00 | 0.98 | 0.99 | 50 |
| Pinhole | 0.99 | 0.98 | 0.99 | 124 |
| Hole | 0.98 | 1.00 | 0.99 | 258 |
| Spatter | 0.99 | 0.98 | 0.99 | 318 |
| Accuracy | - | - | 0.99 | 750 |
| Macro avg | 0.99 | 0.99 | 0.99 | 750 |
| Weighted avg | 0.99 | 0.99 | 0.99 | 750 |

For defect Crack (Class 0), the SSFSL approach exhibited a precision of 1.00, demonstrating that instances predicted as Class 0 were correct 100% of the time. The recall of 0.94 indicates that the model successfully identified 94% of instances with Class 0 defects, and the resulting f1-score of 0.97 emphasizes the balanced performance between precision and recall. With a support of 79 instances, the SSFSL approach excelled in accurately capturing instances of Class 0 defects. Comparatively, in the supervised approach, the precision, recall, and f1-score for defect type 0 were 1.00, 0.98, and 0.99, respectively. The SSFSL approach demonstrated a slightly lower recall but maintained competitive precision and f1-score, highlighting its comparable performance in defect detection for this category.

For defect Pinhole (Class 1), the SSFSL approach achieved precision, recall, and f1-score values of 0.99, 1.00, and 1.00, respectively. These metrics indicate the model's robust ability to accurately identify and classify instances of Class 1 defects, showcasing its reliability in this category. With a support of 107 instances, the SSFSL approach demonstrated excellent accuracy in defect classification. In the supervised approach, the precision, recall, and f1-score for defect type 1 were also high, with values of 0.99, 0.98, and 0.99, respectively. Both the SSFSL and supervised approaches demonstrated comparable performance in defect detection for Class 1.

Defect Hole (Class 2) in the SSFSL approach showcased precision, recall, and f1-score values of 1.00, 0.99, and 0.99, respectively, indicating near-perfect classification of instances for this defect category. The supervised approach, with precision, recall, and f1-score values of 0.98, 1.00, and 0.99, respectively, also demonstrated high accuracy in defect detection for Class 2.

For defect Spatter (Class 3), the SSFSL approach achieved precision, recall, and f1-score values of 0.98, 1.00, and 0.99, respectively. The high precision suggests that the model rarely misclassified instances as Class 3 when they belonged to other categories, while the recall of 1.00 indicates effective identification of all instances of Class 3 defects. With 347 instances, the SSFSL approach exhibited strong performance in capturing instances of this defect type.

In the supervised approach, precision, recall, and f1-score for defect type 3 were 0.99, 0.98, and 0.99, respectively, demonstrating comparable performance to the SSFSL approach. The overall accuracy of 99% across all defect types, as indicated by the macro and weighted averages (*Table 2*), underlines the SSFSL approach's ability to generalize well and perform robustly on the entire dataset. The comparison with the supervised approach highlights the SSFSL approach's efficiency in defect detection, especially in scenarios with limited labeled data, offering a promising avenue for enhancing defect identification in Laser Powder Bed Fusion processes.

Table 2. Semi-Supervised Few-Shot Learning of Unlabeled Images by Pseudo Labelling methodology.

| Label | Precision | Recall | F1-score | Support |
|---|---|---|---|---|
| **Crack** | 1.00 | 0.94 | 0.97 | 79 |
| **Pinhole** | 0.99 | 1.00 | 1.00 | 107 |
| **Hole** | 1.00 | 0.99 | 0.99 | 296 |
| **Spatter** | 0.98 | 1.00 | 0.99 | 347 |
| **Accuracy** | | | 0.99 | 829 |
| **Macro avg** | 0.99 | 0.99 | 0.99 | 829 |
| **Weighted avg** | 0.99 | 0.99 | 0.99 | 829 |

The confusion matrix, depicted in *Figure 10*, showcased the SSFSL model's superiority over the supervised model, with 98.66% accuracy for Crack, 100% for Pinhole, 98.64% for Hole, and 100% for Spatter. Despite the high accuracy, logical errors were identified, such as a crack being mistakenly predicted as Pinhole or Spatter due to overlapping defects, and the simultaneous inclusion of Hole and Spatter due to laser beam instability. These insights inform the need for further refinement in the model to address such challenges.

In conclusion, the integration of semi-supervised learning, transfer learning, pseudo-labeling, and the TransMatch algorithm proved to be an efficient strategy for enhancing the performance of defect detection in LPBF. This approach not only overcame the challenges associated with limited labeled data but also demonstrated its adaptability and effectiveness in a real-world manufacturing scenario. Ongoing refinements and optimizations will undoubtedly further improve the model's capabilities, making it a valuable tool in quality control for LPBF processes.

In the confusion matrix, despite the high overall accuracy achieved by the Semi-Supervised Few-Shot Learning (SSFSL) approach, certain errors and misclassifications highlight potential limitations of the model. Some cracks were misclassified as Pinhole or Spatter, possibly due to overlapping or visually similar features among the defect types. Hole and Spatter defects occasionally appeared together in the model's predictions, suggesting difficulties in distinguishing between defects that may result from laser beam instability and present combined patterns. False negatives also arose when the model relied heavily on unstable visual features influenced by printing process variations. Overlapping defects introduced further challenges, as multiple defect types within the same region demand a more nuanced classification approach. Sensitivity to noise and artifacts in Laser Powder Bed Fusion images underscored the need for robust preprocessing or training adjustments. Additionally, inaccuracies in the pseudo-labels generated during semi-supervised learning could have contributed to certain misclassifications, emphasizing the importance of carefully validating and refining these labels.

Refining model architecture, incorporating advanced feature extraction techniques, and enhancing the model's capacity to handle complex or overlapping defect scenarios offer promising directions for improvement. Iterative refinements guided by confusion matrix insights can strengthen the model's robustness, ultimately leading to a more accurate defect detection system.

Comparison with previous work using the transMatch algorithm for detecting surface defects fabricated by LPBF reveals strong performance from the proposed semi-supervised few-shot learning approach. The model achieves an overall accuracy of 99% on unlabeled images, indicating impressive capability in identifying various defect types. Furthermore, the confusion matrix and classification report demonstrate consistently high precision, recall, and F1-scores across defect categories 0, 1, 2, and 3. This performance underlines the method's effectiveness in accurately discerning and classifying defects within Laser Powder Bed Fusion processes.

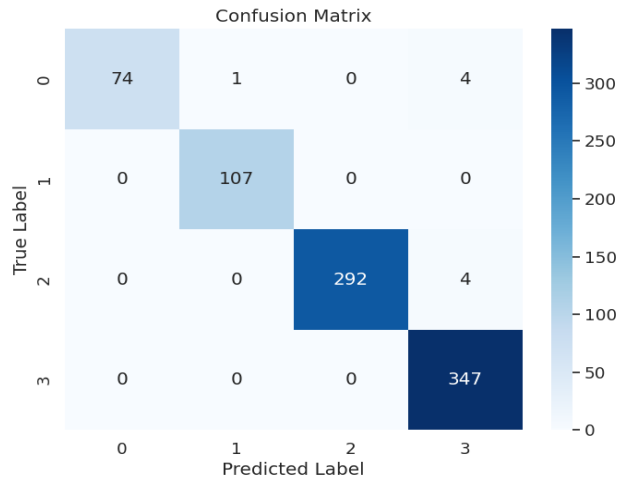

**Figure 10.** Confusion Matrix of TransMatch Algorithm through SSFSL.

Comparing our results with prior works in supervised learning, particularly those utilizing artificial neural networks (ANN) and convolutional neural networks (CNN), in ***Table 3***, our model's accuracy stands out. While some supervised models achieve notable accuracies ranging from 84.45% to 100%, our semi-supervised approach surpasses these benchmarks with an accuracy of 99%. This underscores the efficacy of incorporating transMatch for unlabelled image classification in the context of LPBF surface defect detection.

Furthermore, our approach outperforms existing unsupervised learning models, such as k-Means, which achieves a comparatively lower accuracy of 60% in identifying pores, keyholes, and lack of fusion defects. The use of semi-supervised learning in our model capitalizes on both labelled and unlabelled data, enhancing the robustness and accuracy of defect detection.

In the context of semi-supervised learning, our model's performance aligns with or exceeds that of other approaches. Notably, our accuracy of 99% is comparable to the results achieved by Graph Neural Networks (GNNs) in detecting porosity (99%) and Siamese Neural Networks (SNN) in identifying various defects, including porosity, overheating, balling, lack of fusion (LoF), and spattering (86% and 70%).

Moreover, our model's accuracy in detecting specific defect types, such as balling, lack of fusion (LoF) pores, keyhole pores, delamination, and crack propagation (97%), surpasses the performance of other semi-supervised learning models employing convolutional neural networks (CNN).

In conclusion, our semi-supervised few-shot learning approach, leveraging the TransMatch algorithm, demonstrates superior accuracy in surface defect detection for LPBF-fabricated components compared to both supervised and unsupervised counterparts. This showcases the potential of our methodology in achieving high precision and recall across diverse defect categories, contributing to the advancement of efficient and reliable quality control in additive manufacturing processes.

Table 3. Comparison of Defects Detection Methodologies.

| ML category | ML algorithm | Defect type | Accuracy | Ref. |
|---|---|---|---|---|
| **Supervised Learning** | Artificial Neural Network (ANN) | Anomalous | 90% | [49] |
| | | Porosity | 84.45% | [50] |
| | Convolutional Neural Network (CNN) | Crack, gas porosity, lack of fusion | 92.1% | [51] |
| | | Porosity | 91.2% | [52] |
| | | Porosity | 100% | [53] |
| | | Delamination, splatters, good quality | 96.80% | [2] |
| | | Defects detection in thin walls. | 85-98% | [54] |
| | | Defects detection | 92% | [55] |
| | | Delamination and spatter detection | 97% | [56] |
| **Unsupervised Learning** | k-Means | Pores, keyholes, and lacks of fusion defects | 60% | [57] |
| **Semi-Supervised Learning (SSL)** | Graph neural networks (GNNs) | Porosity | 99% | [58] |
| | Siamese Neural Networks (SNN) | Hierarchical classifier, One-shot classifier Porosity, Overheating, Balling, Lack of fusion (LoF) and Spattering | 86%, 70% | [59] |
| | Convolutional Neural Network (CNN) | Balling, Lack of fusion (LoF) pores, keyhole pores, delamination, and crack propagation | 97% | [60] |
| | **CNN** | **Defect Classifier: Crack, Pinhole, Hole, Spatter** | 98%-99.61% | **This Work** |

## 4. Conclusion

This paper has introduced a novel Semi-Supervised Few-Shot Learning (SSFSL) framework for surface defect detection in LPBF-built Ti alloys, leveraging a CNN pre-trained on 34 labeled images and a multi-iteration pseudo-labeling strategy. By combining transfer learning with SSFSL, the model effectively exploits both labeled and unlabeled data to classify four main defect types—Crack, Pinhole, Hole, and Spatter. The initial supervised phase already demonstrated promising accuracy, but incorporating SSFSL further refined the performance, achieving an overall accuracy of 0.9891 with minimal loss (0.0188). Notably, the model attained accuracy values of 93.67% for Crack, 100% for Pinhole, 98.64% for Hole, and 100% for Spatter, underscoring its robustness even in the face of diverse geometries and overlapping defect features.

This work highlights the advantages of semi-supervised approaches for mitigating the heavy annotation demands typical of additive manufacturing defect datasets. By progressively learning from large volumes of unlabeled images, the proposed SSFSL algorithm demonstrated substantial gains over purely supervised methods, especially for defects prone to misclassification under limited training data. The model's strong classification metrics across multiple defect categories affirm its potential as a versatile and efficient tool for real-world additive manufacturing quality control. With further refinements in architecture and feature extraction, SSFSL-based strategies can continue to drive advancements in automated defect detection, ultimately contributing to improved reliability and cost savings in industrial applications.